\documentclass[runningheads]{llncs}
\usepackage{anyfontsize}
\usepackage[dvipsnames]{xcolor}

\usepackage{tikz}
\usepackage{wrapfig}
\usepackage{comment}
\usepackage{amsmath,amssymb} 

\usepackage[accsupp]{axessibility}  
\usepackage{graphicx}
\usepackage{amsmath}
\usepackage{amssymb}
\usepackage{booktabs}
\usepackage{times}
\usepackage{epsfig}
\usepackage{enumitem}
\usepackage{bbm}
\usepackage{multirow}
\usepackage{caption} 
\usepackage{array}
\usepackage{arydshln}
\usepackage[square,numbers]{natbib}
\usepackage{subcaption}
\usepackage{pifont}

\usepackage[pagebackref=true,breaklinks=true,letterpaper=true,colorlinks,bookmarks=false]{hyperref}
\usepackage[capitalise]{cleveref}                 

\usepackage{listings}
\definecolor{codegreen}{rgb}{0,0.6,0}
\definecolor{codegray}{rgb}{0.5,0.5,0.5}
\definecolor{codepurple}{rgb}{0.58,0,0.82}
\definecolor{backcolour}{rgb}{0.95,0.95,0.92}

\lstdefinestyle{mystyle}{
    backgroundcolor=\color{backcolour},   
    commentstyle=\color{codegreen},
    keywordstyle=\color{magenta},
    numberstyle=\tiny\color{codegray},
    stringstyle=\color{codepurple},
    basicstyle=\ttfamily\footnotesize,
    breakatwhitespace=false,         
    breaklines=true,                 
    captionpos=b,                    
    keepspaces=true,                 
    numbers=left,                    
    numbersep=5pt,                  
    showspaces=false,                
    showstringspaces=false,
    showtabs=false,                  
    tabsize=2
}

\makeatletter
\newcommand{\printfnsymbol}[1]{%
  \textsuperscript{\@fnsymbol{#1}}%
}

\makeatletter
\newcommand{\dummylabel}[2]{\def\@currentlabel{#2}\label{#1}}
\makeatother

\begin{document}
\pagestyle{headings}
\mainmatter
\def\ECCVSubNumber{5536}  

\title{TextAdaIN: Paying Attention to Shortcut Learning in Text Recognizers} 



\titlerunning{TextAdaIN: Paying Attention to Shortcut Learning in Text Recognizers}
%
\author{Oren Nuriel \and
Sharon Fogel \and
Ron Litman}
\authorrunning{Nuriel et al.}
%
\institute{AWS AI Labs\\
\email{\{onuriel, shafog, litmanr\}@amazon.com}}


\newcommand{\etal}{\textit{et al.}}
\newcommand{\AlgoName}{TextAdaIN }
\newcommand{\AlgoNameNoSpace}{TextAdaIN}

\maketitle

\begin{abstract}
Leveraging the characteristics of convolutional layers, neural networks are extremely effective for pattern recognition tasks. However in some cases, their decisions are based on unintended information leading to high performance on standard benchmarks but also to a lack of generalization to challenging testing conditions and unintuitive failures. Recent work has termed this "shortcut learning" and addressed its presence in multiple domains. In text recognition, we reveal another such shortcut, whereby recognizers overly depend on local image statistics. Motivated by this, we suggest an approach to regulate the reliance on local statistics that improves text recognition performance.

Our method, termed \AlgoNameNoSpace, creates local distortions in the feature map which prevent the network from overfitting to local statistics. It does so by viewing each feature map as a sequence of elements and deliberately mismatching fine-grained feature statistics between elements in a mini-batch.
Despite \AlgoNameNoSpace's simplicity, extensive experiments show its effectiveness compared to other, more complicated methods. 
\AlgoName achieves state-of-the-art results on standard handwritten text recognition benchmarks. It generalizes to multiple architectures and to the domain of scene text recognition. Furthermore, we demonstrate that integrating \AlgoName improves robustness towards more challenging testing conditions. The official Pytorch implementation can be found at \url{https://github.com/amazon-research/textadain-robust-recognition}.
\end{abstract}

\section{Introduction}

Reading someone else's handwriting is often a challenging task; some of the characters are unclear, the text is cursive, there is background clutter and the image quality can be low. When deciphering each character, we often rely on the surrounding area to compensate for the occasional obscurity of the text. The automation of reading text images has been a thriving field of research in computer vision for decades. Recent deep learning methods have significantly improved recognition results \cite{Baek2019clova,zhang2019sequence,luo2020learn,Litman_2020_CVPR,aberdam2020sequence,aberdam2022multimodal}.

\begin{figure}[t]
  \centering
  \includegraphics[width=0.7\columnwidth]{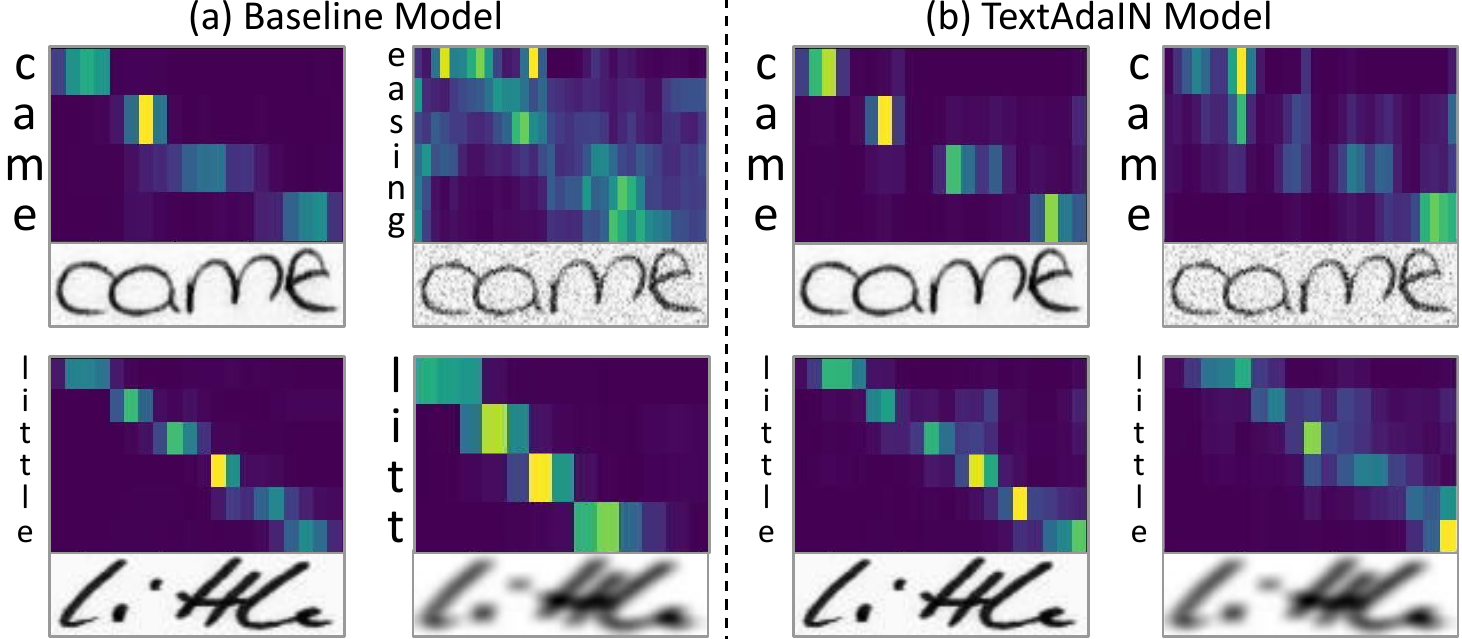}

  \caption{\textbf{Decoder attention maps.} Each example shows the input image (bottom), attention map (top) and model prediction (left). Each line in the attention map is a time step representing the attention per character prediction. (a) The baseline model, which uses local statistics as a shortcut, misinterprets the corrupted images. (b) Our proposed method which overcomes this shortcut, enhances performance on both standard and challenging testing conditions}
  \label{fig:attn}
    \vspace{-0.4cm}
\end{figure}

However, previous works suggest that despite their super-human capabilities, deep learning methods are limited by their tendency to err even when introducing small (in some cases even invisible) modifications to the input. Numerous works have touched on this subject from various angles \cite{bickel2009discriminative, scholkopf2012causal, torralba2011unbiased, szegedy2013intriguing, geirhos2018, nuriel2021permuted} and propose tailored solutions. Geirhos \etal~\cite{geirhos2020shortcut} view each of these as a symptom of the same underlying problem, and terms this phenomenon \textit{shortcut learning - decision rules that perform well on independent and identically distributed (i.i.d.) test data but fail on tests that are out-of-distribution (o.o.d)}.
Shortcut learning occurs when there is an easy way to learn attributes which are highly correlated with the label (\emph{i.e.,} ``it stands on a grass lawn $\implies$ it is a cow''). Thus, shortcuts are inherent in the data \cite{geirhos2018, hermann2019origins}.
Yet, as they can depend on other various factors such as the architecture \cite{d2019finding}, and the optimization procedure \cite{wu2017towards, de2018deep} as well, revealing an instance of shortcut learning is  non-trivial task.
The fact that a model learned a shortcut can be discovered at test time when it encounters examples that the shortcut heuristic fails on (\emph{i.e.,} a cow on the beach).

In this work, we reveal a shortcut pertaining to text recognizers, specifically, we reveal the unhealthy reliance of text recognizers on local statistics. Based on the observation that text recognizers operate on a local level, we hypothesize that they are susceptible to overly rely on local information which may even be indistinguishable to the naked eye (\emph{i.e.,} a certain level of curvature is unique and highly correlated with the character ``f''). As such in this case it is difficult to identify the exact shortcut heuristic, thus we provide intuition through quantitative and qualitative analyses. \cref{fig:attn}(a)  illustrates the decoder's attention maps for a state-of-the-art \cite{Litman_2020_CVPR} recognizer before and after applying local corruptions to the image. In the first row, additive Gaussian noise is applied, which distorts local information while maintaining semantic information. As a consequence, the model diverges completely. In the second row, motion blur is applied, artificially imitating object motion. In this case, in order to successfully decode the corrupted image, global context is necessary (imagine reading the blurred image character by character without access to the surrounding characters). As one can tell from the attention maps, the model fails to do so and is unable to correctly decode the image. An in-depth analysis reveals that this phenomenon is apparent in standard text recognizers across both scene text and handwritten domains, as further elaborated in \cref{sec:method}.

\begin{figure}[t]
  \centering
  \includegraphics[width=0.85\columnwidth]{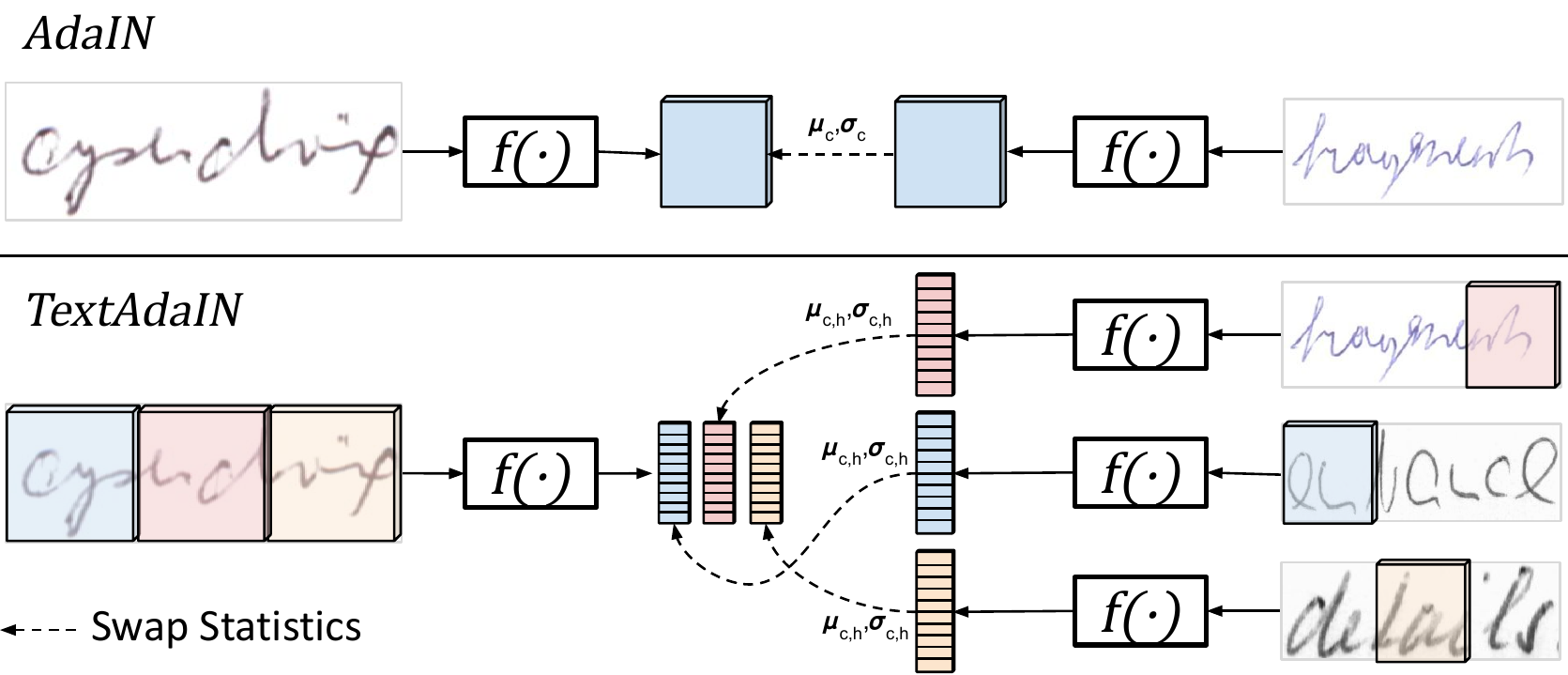}

  \caption{\textbf{\AlgoNameNoSpace.} \AlgoName views each of the feature maps as a sequence of individual elements and swaps feature statistics between elements instead of entire images as in AdaIN. This feature-level distortion alleviates the model's tendency to rely on shortcuts in the form of local statistics}
  \label{fig:teaser_fig}
 \vspace{-0.4cm}
\end{figure}
To prevent the model from using the aforementioned shortcut, we propose a simple, yet powerful, technique for moderating the reliance on local statistics. The key idea of our approach is to probabilistically swap fine-grained feature statistics in a manner that is adjusted to the task at hand.
To that end, we propose \AlgoNameNoSpace, a local variation of AdaIN~\cite{huang2017arbitrary}, depicted in \cref{fig:teaser_fig}. In contrast to AdaIN which operates on the entire image, \AlgoName splits the feature map into a sequence of elements and operates on each element independently. Furthermore, the normalization is performed over multiple dimensions. Both modifications increase the granularity level in which the statistics are modified and enable the usage of multiple donor images. Effectively, the representation space undergoes distortions derived from an induced distribution, namely other text images, at a sub-word level. Thus, forcing the encoder to account for information in surrounding areas as the local information cannot always be depended on. This is observed in \cref{fig:attn}(b), where in contrast to the baseline model, \AlgoName successfully utilizes global context and properly decodes the images despite the corruptions.

We validate our method by comparing its performance with state-of-the-art approaches on several handwritten text benchmarks.
\AlgoName achieves state-of-the-art results, reducing the word error rate by $\textbf{11.8\%}$ and $\textbf{16.4\%}$ on IAM and RIMES, respectively. Furthermore, our method shows a consistent improvement across multiple architectures and in the domain of scene text recognition.
Not only does our model surpass other, more complicated methods (\cite{bhunia2019handwriting, fogel2020scrabblegan, zhang2019sequence, luo2020learn, aberdam2020sequence}), but it is simple to implement and can effortlessly be integrated into any mini-batch training procedure.
To summarize, the key contributions of our work are:
\begin{enumerate}[nolistsep,leftmargin=*]
\item We reveal an instance of shortcut learning in text recognizers in the form of a heavy dependence on local statistics and suggest to regulate it.
\item We introduce \AlgoNameNoSpace, a simple yet effective normalization approach to remediate the reliance on local statistics in text recognizers.
\item Extensive experimental validation shows our method achieves state-of-the-art results on several popular handwritten text benchmarks. In addition, it is applicable to the domain of scene text and can be used independent of the chosen architecture.
\item We demonstrate that our method leads to improved performance on challenging testing conditions. 
\end{enumerate}

\section{Related Work}

\paragraph{\textbf{Shortcut learning.}}
An extensive amount of research has been conducted at understanding the behaviour of neural networks. One behavioural aspect under investigation is their unintended solutions, whereby a decision is made based on misleading information, thus limiting the performance under general conditions. A typical example of this is an intriguing discovery made by Szegedy \etal~\cite{szegedy2013intriguing}. They showed the susceptibility of neural networks to adversarial examples, which are samples that undergo minimal, or even unnoticeable,  modifications capable of altering model predictions. More recent works show other ill-desired sensitivities, \cite{geirhos2018} for texture, \cite{afifi2019else} for color constancy, \cite{nuriel2021permuted, Sakai-1971-15077} for global statistics and \cite{beery2018recognition} for image background. These phenomena were classified by Geirhos \etal~\cite{geirhos2020shortcut} as instances of the same underlying problem that they termed \textit{shortcut learning}. So far, for shortcut learning in the domain of text recognition, we are only aware of \cite{wan2020vocabulary}, which exposed the decoder's high reliance on vocabulary.

\paragraph{\textbf{Normalization and style transfer.}}
Normalizing feature tensors is an effective and powerful technique that has been explored and developed over the years for a variety of tasks. Ioffe \& Szegedy~\cite{ioffe2015batch} were the first to introduce Batch Normalization, which inspired a series of normalization-based methods such as Layer Normalization \cite{ba2016layer}, Instance Normalization \cite{ulyanov2016instance}, and Group Normalization \cite{wu2018group}.

The style of an image was characterized by \cite{gatys2016image} to be the statistics of activation maps in intermediate layers of convolutional neural networks. Instance Normalization (IN) \cite{ulyanov2016instance} is a normalization layer which normalizes these statistics, removing the style representation from the activation map. Subsequently, Adaptive Instance Normalization (AdaIN) was proposed by \cite{huang2017arbitrary} for real-time arbitrary style transfer. AdaIN changes the style of an image by incorporating statistics from an additional image. 

Leveraging the benefits of this operation, Geirhos \etal~\cite{geirhos2018} created Stylized-ImageNet, a stylized version of ImageNet. They demonstrated that classifiers trained on this dataset rely more on shape than on texture.
More recently, Zhou \etal~\cite{zhou2021mixstyle} proposed to probabilistically mix instance-level feature statistics during training across different domains, thus increasing the generalizability of the model. 
Furthermore, Nuriel~\etal~\cite{nuriel2021permuted} demonstrated that a similar approach can reduce classifiers' reliance on global statistics, therefore increasing classification performance.

In this work we reveal a shortcut learning phenomenon in text recognizers. To remediate this problem, we also leverage AdaIN. However, instead of operating on a word (instance) level, we swap fine-grained statistics on a sub-word level. We do so by treating every few consecutive frames in the sequential feature map as individual elements.

\paragraph{\textbf{Text Recognition.}}
Text recognition has attracted considerable attention over the past few years. In particular, deep learning approaches have achieved remarkable results \cite{shi2016end,Baek2019clova,Litman_2020_CVPR,slossberg2020calibration, fang2021read, wang2021two,aberdam2022multimodal}. 
Still, current state-of-the-art methods struggle to train robust models when the amount of data is insufficient to capture the magnitude of styles.
Various methods have been suggested to cope with this problem. Bhunia \etal~\cite{bhunia2019handwriting} proposed an adversarial feature deformation module that learns ways to elastically warp extracted features, boosting the network's capability to learn highly informative features. Luo \etal~\cite{luo2020learn} introduced an agent network that learns from the output of the recognizer. The agent controls a set of fiducial points on the image and uses a spatial transformer to generate “harder” training samples.

Unlike previous methods, our method does not require additional data, new models or complex training paradigms.
Instead, we suggest a normalization-based method adjusted to text images and to sequence-to-sequence approaches.
\AlgoName is extremely easy to implement and can fit into any encoder as part of a mini-batch training process.

Throughout this work, unless mentioned otherwise, we integrate our proposed method with a state-of-the-art recognizer named SCATTER~\cite{Litman_2020_CVPR}. 
The SCATTER architecture consists of four main components: spatial transformation, feature extraction, visual feature refinement and a selective-contextual refinement block.
For further details about the baseline architecture, we refer the reader to \cref{app:scatter}.

\section{Method}
\label{sec:method}

\subsection{Shortcut Learning in Text Recognizers}
\label{sec:shortcut learning}
\setlength{\intextsep}{-0.4cm}%
\begin{wraptable}[19]{r}{8cm}
\small
\begin{center}

\bgroup
  \captionsetup[table]{skip=5pt}
    \caption{ \textbf{Text recognizers' performance in challenging testing conditions.} Mean accuracies of different state-of-the-art text recognizers while applying a series of subtle local image corruptions. The corruptions consist of local masking and pixel-wise distortions: Cutout, Dropout, Additive Guassian Noise, Elastic Transform and Motion Blur. Models are especially susceptible to learning shortcuts when the training data is limited, as in handwriting}
\centering

  \begin{center}
    \small
    \resizebox{8cm}{!}{
    \begin{tabular}{llll} 
      \toprule
      \multirow{2}{*}{\textbf{Method}} & \multicolumn{2}{c}{\textbf{Scene Text}} & \multicolumn{1}{c}{\textbf{Handwritten}}
      \\  & \multicolumn{1}{c}{Regular text} & \multicolumn{1}{c}{Irregular text} & \multicolumn{1}{c}{IAM} \\
      \midrule
      Baek et al. (CTC) \cite{Baek2019clova} & 88.7 & 72.9 & 80.6   \\
      +Local Corruption & 69.8 \color{Red}{(\textbf{-18.9})} & 44.1 \color{Red}{(\textbf{-28.8})} & 40.4 \color{Red}{(\textbf{-40.2})}  \\
      \midrule
      Baek et al. (Attn) \cite{Baek2019clova} & 92.0 & 77.4 & 82.7 \\
      +Local Corruption & 74.5 \color{Red}{(\textbf{-17.5})} & 50.5 \color{Red}{(\textbf{-26.9})} & 46.2 \color{Red}{(\textbf{-36.5})}  \\
      \midrule
      SCATTER \cite{Litman_2020_CVPR} & 93.5 & 82.1  & 85.7 \\
      +Local Corruption & 76.8 \color{Red}{(\textbf{-16.7})} & 55.3 \color{Red}{(\textbf{-26.8})} & 54.8 \color{Red}{(\textbf{-30.9})}  \\
      \bottomrule
    \end{tabular}}
    \small

    \label{tab:curr_small}
  \end{center}
\egroup
\end{center}
\end{wraptable}




As elaborated on in \cite{geirhos2020shortcut}, shortcuts are dependent on various factors, including the architecture of the model and the data. We therefore draw our motivation for suspecting local statistics as a potential instance of shortcut learning in text recognizers based on two key observations pertaining to both the architecture and the nature of the data. First, the characters appear across a series of frames along the width axis of the image. Therefore, text recognition approaches define the recognition task as a sequence character classification problem.
This leads us to the second observation, the model's predictions are mostly based on local information which lies in consecutive frames~\cite{Bai2015crnn,qiao2020seed}. Local information is referred to, in our work, as local feature statistics. 

As illustrated in Figure \ref{fig:attn}, when applying slight local modifications to the image distorting local information in a way which does not affect human reading capabilities (see \cref{app:corruption}), the baseline model is unable to predict the words correctly. Table \ref{tab:curr_small} quantitatively exemplifies this phenomenon, whereby the models fail to generalize to images under challenging testing conditions in the form of local corruptions.
The table shows the performance of three different text recognition architectures - Baek et al. \cite{Baek2019clova} with CTC and attention heads and SCATTER \cite{Litman_2020_CVPR}. 
All models have been tested on the original scene-text and IAM datasets and on corrupted versions of these datasets. The corruptions used are the local-masking and pixel-wise distortions described in Table \ref{tab:corruption}. All of the models suffer from a large decline in performance when adding the corruptions, and especially on handwriting where the training data is scarce.

An additional experiment validating text recognizers' sensitivity to local distortions is shown in \cref{sec:reliance}, where we empirically demonstrate that text recognizers heavily rely on local information. 
Specifically, we show that profusely distorting global statistics has little to no effect on performance (unlike in classification models~\cite{nuriel2021permuted}). However, intensely distorting local statistics substantially degrades performance, which leads to the understanding that text recognizers do indeed rely on such information.  \setlength{\intextsep}{0cm}%
\begin{wrapfigure}[25]{R}{7cm}

  \centering

  \includegraphics[width=7.5cm]{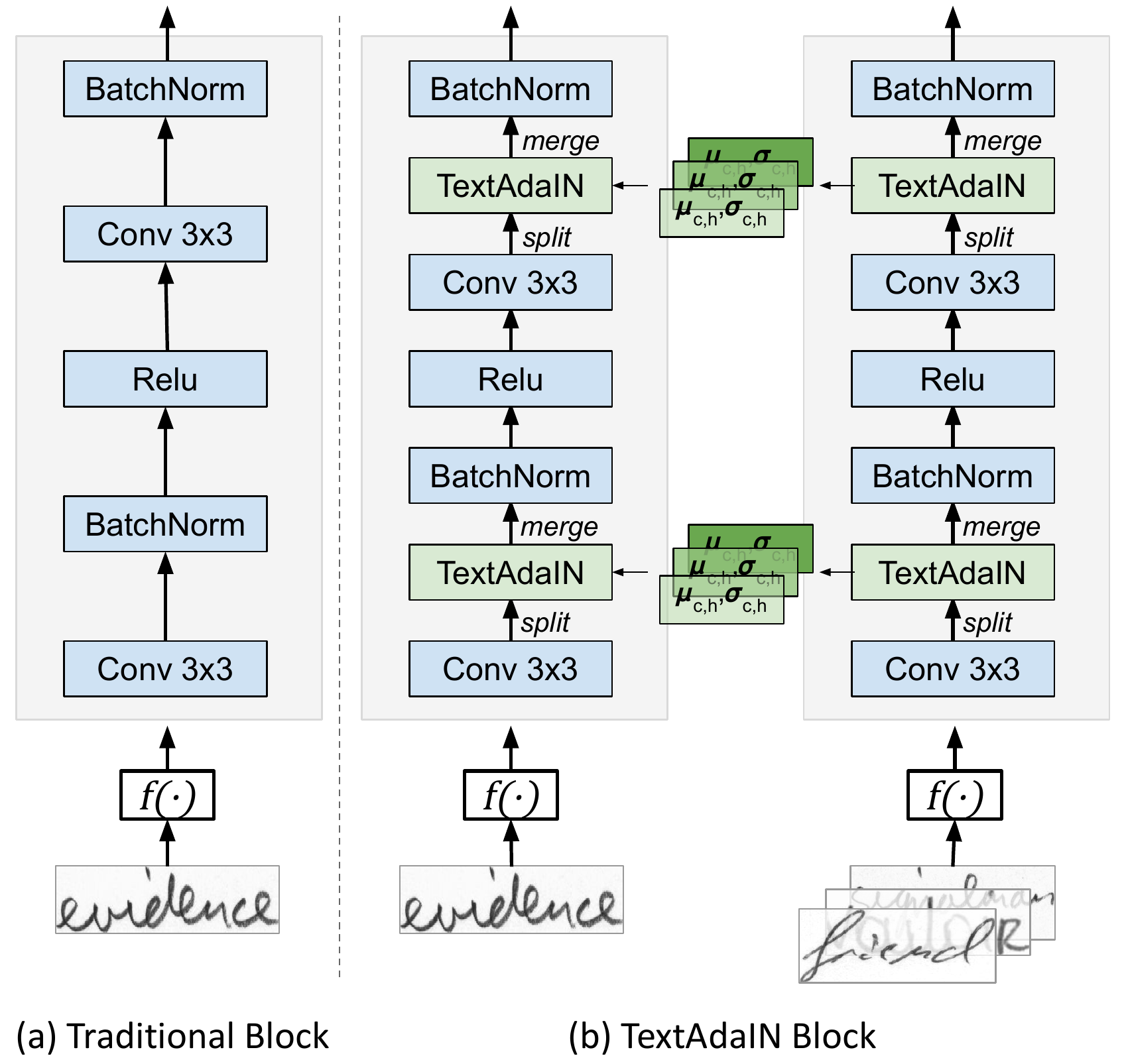}
   \caption{{\bf \AlgoName block.} (a) A standard residual block and (b) a \AlgoName block. \AlgoName is probabilistically employed after every Conv layer during training. \textit{split} and \textit{merge} refer to the mapping operations of reshaping each feature map and $\mu_{c,h}, \sigma_{c,h}$ refer to the feature statistics, as described in \cref{sec:TextAdaIN}}
  \label{fig:arch_fig}
\end{wrapfigure}
The local statistics are an important cue, although they are not a determining factor. Thus, to regulate this ill-desired model bias, we propose \AlgoNameNoSpace, a sequence-aware local variation of AdaIN.

\subsection{AdaIN}
To define AdaIN, we first begin with formally describing Instance Normalization (IN) \cite{ulyanov2016instance}. Given an instance $x\in \mathbb{R}^{C\times H \times W} $, where $C,H$ and $W$ are the channels, height, and width respectively, IN is defined to be:
\begin{equation}
    \textrm{IN}(x)= \gamma\left(\frac{x-\mu(x)}{\sigma(x)}\right)+\beta\,.
\end{equation}
Where $\gamma, \beta \in \mathbb{R}^{C}$ are learned parameters and $\mu(x), \sigma(x) \in \mathbb{R}^{C}$ are calculated by taking the mean and standard deviation over $H,W$ of $x$. 

Adaptive Instance Normalization (AdaIN), proposed in~\cite{huang2017arbitrary}, is built upon IN. Given two image representations, $x_a,x_b\in \mathbb{R}^{C \times H \times W}$, AdaIN shifts the statistics of the representation of $x_a$ to the representation of $x_b$. This is done in two steps. First, Instance Normalization is applied on $x_a$ to remove $x_a$'s style information. Then, the normalized activation map is scaled and shifted to match $x_b$'s statistics. This operation is perceived to transfer $x_b$'s style onto $x_a$.
\begin{equation}
  \label{eq:2}
\textrm{AdaIN}_c(x_a,x_b)= \sigma(x_b)\left(\frac{x_a-\mu(x_a)}{\sigma(x_a)}\right)+\mu(x_b)\,.
\end{equation}
$\textrm{AdaIN}_c$ denotes the standard AdaIN operation in which $\sigma,\mu$ are calculated over the spatial dimensions, resulting in shifting corresponding channel statistics.
\subsection{TextAdaIN}
\label{sec:TextAdaIN}

We wish to design a method that produces model decision rules which are not overly reliant on local information. To this end, we propose a method which distorts local feature statistics during training. Leveraging the recent development proposed by \cite{nuriel2021permuted, zhou2021mixstyle},
we suggest two modifications to the vanilla AdaIN operation that correspond to the aforementioned observations:
\begin{enumerate}[nolistsep]
\item Viewing the feature map as a sequence of individual elements and swapping the feature statistics between elements instead of entire images.
\item Modifying AdaIN to operate on two dimensions - the height and channels of the feature map.
\end{enumerate}
These modifications increase the granularity level in which statistics are calculated and modified, thus regulating the reliance on local statistics. In addition, the sequential view enables the utilization of multiple donor images. Both are crucial for our method's success, as shown in \cref{sec:ablation}.

Formally, given an image representation $x\in \mathbb{R}^{C\times H\times W}$, we define $\mu_{c,h}$, $\sigma_{c,h}$ to be the following:
\begin{equation}
    \mu_{c,h}(x) = \frac{1}{W}\sum_{w=1}^{W} x_{c,h,w} \,,
\end{equation}
\begin{equation}
    \sigma_{c,h}(x) = \sqrt{\frac{1}{W}\sum_{w=1}^{W}(x_{c,h,w} - \mu_{c,h}(x))^{2} + \epsilon}\ \,.
\end{equation}
Therefore, a local variation of $\textrm{AdaIN}_c$ can be defined as:
\begin{equation}
    \textrm{AdaIN}_{c,h}(x_a,x_b)= \sigma_{c,h}(x_b)\left(\frac{x_a-\mu_{c,h}(x_a)}{\sigma_{c,h}(x_a)}\right)+\mu_{c,h}(x_b) \,.
\end{equation}
This variant of AdaIN swaps statistics for every corresponding channel and height, thus impacting the feature map's statistics at a higher level of granularity. We note that backpropagating gradients only occur through $\mu_{c,h}(x_a), \sigma_{c,h}(x_a)$. The reason behind this is to avoid gradient flow from input image labels to the donor images.
\begin{table*}
\normalsize
\begin{center}
\small
\bgroup
\captionsetup[table]{skip=5pt}
\caption{\textbf{Comparison to previous methods.} Word and character error rates (WER and CER) are measured on IAM, RIMES and CVL handwriting datasets. '*' indicates using the unlabeled test for training. Our method achieves state-of-the-art across all datasets}
\label{tab:htr_sota_results}
\def\arraystretch{1.1}
\centering
  \begin{center}
    \small
    \begin{tabular}{lccc|ccc|ccc} 
      \toprule
      \multirow{2}{*}{\textbf{Method}} & \multicolumn{3}{c|}{\textbf{IAM}} & \multicolumn{3}{c|}{\textbf{RIMES}} & \multicolumn{3}{c}{\textbf{CVL}}\\
      & WER & CER & Average & WER & CER & Average & WER & CER & Average \\
      \midrule
      Bluche et al.~\cite{bluche2015deep} & 24.7 & 7.3 & 16.00 & - & - & - & - & - & - \\
      Bluche et al.~\cite{bluche2016joint} & 24.6 & 7.9 & 16.25 & - & - & - & - & - & - \\
      Sueiras et al.~\cite{sueiras2018offline} & 23.8 & 8.8 & 16.30 & - & - & - & - & - & - \\
      Alonso et al.~\cite{alonso2019adversarial} & - & - & - & 11.9 & 4.0 & 7.95 & - & - & - \\
      ScrabbleGAN~\cite{fogel2020scrabblegan} & 23.6 & - & -  & 11.3 & - & - & 22.9 & - & -  \\
      SSDAN*~\cite{zhang2019sequence} & 22.2 & 8.5 & 15.35 & - & - & - & - & - & - \\
      Bhunia et al.~\cite{bhunia2019handwriting} & 17.2 & 8.4 & 12.80 & 10.5 & 6.4 & 8.45 & - & - & -  \\
      Kang et al.*~\cite{kang2020unsupervised} & 17.3 & 6.8 & 12.05 & - & - & - & - & - & - \\
      SeqCLR~\cite{aberdam2020sequence} & 20.1 & 9.5 & 14.80 & 7.6 & 2.6 & 5.10 & 22.2 & - & - \\
      Luo et al.~\cite{luo2020learn} & 14.1 & \textbf{5.4} & 9.75 & 8.7 & 2.4 & 5.50 & - & - & - \\
      \midrule
      SCATTER \cite{Litman_2020_CVPR} & 14.4 & 6.4 & 10.40 & 6.7 & 2.3 & 4.50 & 22.3 & 18.9& 20.6 \\
      +MixStyle~\cite{zhou2021mixstyle} & 14.3 & 6.3 & 10.30 & 6.8 & 2.4 & 4.60 & 22.3 & 18.5 & 20.40 \\
      +pAdaIN~\cite{nuriel2021permuted} & 14.6 & 6.4 & 10.50 & 6.4 & 2.2 & 4.30 & 22.3 & 18.2& 20.25 \\
      +\AlgoName & \textbf{12.7} & 5.8 & \textbf{9.25} & \textbf{5.6} & \textbf{1.9} & \textbf{3.70} & \textbf{21.8} & \textbf{18.0} & \textbf{19.90} \\
      \bottomrule
    \end{tabular}
  \end{center}
\egroup
\small

\end{center}
\end{table*}
Given the definitions above, we formulate the \AlgoName operation used during training. Let $X=\{x_i\}_{i=1}^B$ denote a mini-batch of $B$ feature maps. We divide each sample $x_i$ into $K$ windows along its width. The result is a batch of elements pertaining to $B\cdot K$ windows. This operation can be defined as a mapping:
\begin{equation}
 X\in \mathbb{R}^{B\times C \times H \times W} \rightarrow \widehat{X}\in \mathbb{R}^{B\cdot K \times C \times H \times \frac{W}{K}} \,.
 \end{equation}
Then, we employ a similar procedure to the one used in~\cite{nuriel2021permuted}. We randomly draw a permutation of the modified batch $\widehat{X}$ and apply $\textrm{AdaIN}_{c,h}$ between the modified batch and the permuted one. Namely, let 
\begin{equation}
    \pi(\widehat{X}) = [\hat{x}_{\pi(1)},\hat{x}_{\pi(2)}...\hat{x}_{\pi(B\cdot K)}] \,
\end{equation} denote applying a permutation $\pi: [B\cdot K] \rightarrow [B\cdot K]$ on $\widehat{X}$. Then the output of \AlgoName on the $i^{th}$ window of $\widehat{X}$ is defined by:
\begin{equation}
\textrm{\AlgoName}(\hat{x}_i) = \textrm{AdaIN}_{c,h}(\hat{x}_i,\hat{x}_{\pi(i)}) \,.
\end{equation}
Subsequently, the batch of windows is rearranged back to its original form using the inverse mapping operation.

\AlgoName is applied batch-wise with probability $p$ after every convolutional layer in the encoder, as illustrated in \cref{fig:arch_fig}(b). The permutation $\pi$ is sampled uniformly, and $p$ is a hyperparameter fixed ahead of training. \AlgoName is only applied during training and not during inference.


\section{Experiments}

\label{sec:exp}
In this section, we begin by comparing our method's performance against state-of-the-art methods on several public handwriting datasets. Then, we demonstrate that \AlgoName can be integrated into additional recognition architectures and applied to natural scene text images.

\noindent
\textbf{Datasets}~ We conduct experiments on several public handwriting and scene text datasets.
For handwriting, we consider the English datasets IAM~\cite{marti2002iam} and CVL~\cite{kleber2013cvl}, and the French dataset RIMES~\cite{grosicki2009icdar}. For scene text, we train on the synthetic datasets SynthText~\cite{Zisserman206st} and MJSynth~\cite{Zisserman2014mj} and test on four real-world regular text datasets: IIT5K~\cite{Mishra2012sj}, SVT~\cite{Wang2011bottom}, IC03~\cite{Lucas2003ic03}, IC13~\cite{Karatzas2013ic13} and three real-world irregular text datasets: ICDAR2015~\cite{Karatzas2015ic15}, SVTP~\cite{Phan2013svtp} and CUTE 80~\cite{Risnumawan2014cute}.
We present samples from each dataset and include more details in \cref{app:dtatsets}.

\noindent
\textbf{Metrics}~ To evaluate recognition performance, word-level accuracy is measured. For handwritten text recognition state-of-the-art comparison, Word Error Rate (WER) and Character Error Rate (CER) are adopted, similar to the convention used in \cite{sueiras2018offline, zhang2019sequence, aberdam2020sequence}.

\noindent
\textbf{Implementation Details}~ Unless mentioned otherwise, in all of our experiments, \AlgoName is fused into the backbone of SCATTER~\cite{Litman_2020_CVPR}. The experimental settings, including the optimizer, learning rate, image size, and training datasets, are identical to SCATTER~\cite{Litman_2020_CVPR}. Full implementation details are described in \cref{app:implementation_det}.

\subsection{Comparison to State-of-the-Art}
In \cref{tab:htr_sota_results}, we measure the accuracy of our proposed method on public handwritten text benchmarks.
Our method achieves state-of-the-art results across all datasets. Compared to current state-of-the-art methods, incorporating \AlgoName achieves a performance increase of \textbf{+1.4} pp (85.9\% vs. 87.3\%) on IAM, \textbf{+2.0} pp (92.4\% vs. 94.4\%) on RIMES and \textbf{+0.4} pp (77.8\% vs. 78.2\%) on CVL. We wish to emphasize that previous methods, such as \cite{bhunia2019handwriting,kang2020unsupervised,luo2020learn,aberdam2020sequence}, introduced complex modifications to the training phase, including adversarial learning and contrastive pre-training. In contrast, \AlgoName can be easily implemented in a few lines of code and seamlessly fit into any mini-batch training procedure. In addition, the effect of applying MixStyle~\cite{zhou2021mixstyle} and pAdaIN~\cite{nuriel2021permuted} is displayed in \cref{tab:htr_sota_results}. Both have little to no effect on the results, indicating that handwritten text recognizers are already invariant to changes in global statistics.
We refer the reader to \cref{app:fail_cases} to see failure cases of the model.

\begin{table}
\setlength\tabcolsep{3pt}
\begin{center}

\bgroup
\centering
\captionsetup[table]{skip=5pt}
    \caption{ \textbf{Generalization to new domains and architectures.} Similar to \cite{Litman_2020_CVPR}, we show weighted (by size) average results on the regular and irregular scene text datasets. All experiments are reproduced. Integrating \AlgoName results in consistent improvements}
  \begin{center}
    \begin{tabular}{lllll} 
      \toprule
      \multirow{3}{*}{\textbf{Method}} & \multicolumn{2}{c}{\textbf{Scene Text}} & \multicolumn{2}{c}{\textbf{Handwritten}}
      \\  & \multicolumn{1}{c}{Regular text} & \multicolumn{1}{c}{Irregular text} & \multicolumn{1}{c}{IAM} & \multicolumn{1}{c}{RIMES} \\
      & \multicolumn{1}{c}{5,529} & \multicolumn{1}{c}{3,010} & \multicolumn{1}{c}{17,990} &  \multicolumn{1}{c}{7,734} \\
      \midrule
      Baek et al. (CTC) \cite{Baek2019clova} & 88.7 & 72.9 & 80.6 & 87.8  \\
      +\AlgoName & 89.5 \color{OliveGreen}{(\textbf{+0.8})} & 73.8 \color{OliveGreen}{(\textbf{+0.9})} & 81.5 \color{OliveGreen}{(\textbf{+0.9})} & 90.7 \color{OliveGreen}{(\textbf{+2.9})} \\
      \midrule
      Baek et al. (Attn) \cite{Baek2019clova} & 92.0 & 77.4 & 82.7 & 90.2 \\
      +\AlgoName & 92.2 \color{OliveGreen}{(\textbf{+0.2})} & 77.7 \color{OliveGreen}{(\textbf{+0.3})} & 84.1 \color{OliveGreen}{(\textbf{+1.4})} & 93.0 \color{OliveGreen}{(\textbf{+2.8})} \\
      \midrule
      SCATTER \cite{Litman_2020_CVPR} & 93.6 & 83.0  & 85.7 & 93.3 \\
      +\AlgoName & 94.2 \color{OliveGreen}{(\textbf{+0.6})} & 83.4 \color{OliveGreen}{(\textbf{+0.4})} & 87.3 \color{OliveGreen}{(\textbf{+1.6})} & 94.4 \color{OliveGreen}{(\textbf{+1.1})} \\
    \midrule
      AbiNet \cite{fang2021read}  & 93.9 & 82.0  & 85.4 & 92.0 \\
      +\AlgoName & 94.2 \color{OliveGreen}{(\textbf{+0.3})} & 82.8 \color{OliveGreen}{(\textbf{+0.8})} & 86.3 \color{OliveGreen}{(\textbf{+0.9})} & 93.0 \color{OliveGreen}{(\textbf{+1.0})} \\
      \bottomrule
    \end{tabular}

    \label{tab:str_sota_results_avg}
  \end{center}
\egroup
\end{center}
\end{table}

\subsection{Generalization of Proposed Method}
In this subsection, we explore our method's transferability to both the domain of scene text and to different recognition architectures. In addition to the SCATTER and AbiNet \cite{fang2021read} architectures, we utilize the Baek \etal~\cite{Baek2019clova} framework, which can describe the building blocks of many text recognizers, including~\cite{Bai2015crnn,Zisserman2015large,shi2016end,shi2016robust, liu2016star, Hu2017grccn,bai2017accurate,cheng2017focusing, zhang2019sequence, Litman_2020_CVPR, fogel2020scrabblegan, yousef2020origaminet}. 
We choose to present \AlgoNameNoSpace's performance when integrated into Baek \etal~\cite{Baek2019clova} framework while employing either a CTC~\cite{graves2006connectionist} or an attention decoder~\cite{cheng2017focusing, shi2016robust}. As in \cite{Litman_2020_CVPR}, weighted (by size) average word accuracy is adopted where regular and irregular text datasets are distinguished. 

In \cref{tab:str_sota_results_avg}, we present the reproduced performance of the above methods. \AlgoName shows consistent improvement on both scene and handwritten text benchmarks independent of the chosen architecture. The results are competitive with recent state-of-the-art methods.

\begin{wraptable}[22]{R}{7cm}

\small
\bgroup
\def\arraystretch{1.1}
\caption{\textbf{AdaIN dimensions.} Accuracy on the IAM dataset when applying the AdaIN operation over different dimensions. Our method, \AlgoNameNoSpace, shows significant improvement upon the baseline of \textbf{+1.6} pp}

\label{tab:spAdaIN_configuration}
    \begin{tabular}{cccc} 
      \toprule
      Method & AdaIN Dim & Windows & IAM Accuracy \\
      \midrule
      Baseline & - & - & 85.7 \\
      AdaIN  & C & - & 85.4 \\
      \cdashline{1-4}
      \multirow{10}{*}{AdaIN Variations} &
      W & - & 85.7 \\
      &H & - & 85.6  \\
      &H,W & - & 85.7 \\
      &C,W & - & 85.8  \\
      &C,H & - & 85.9 \\
      \cdashline{2-4}
      &C & 5 & 85.9 \\
      &W & 5 & 85.5 \\
      &H & 5 & 85.9  \\
      &H,W & 5 & 85.8 \\
      &C,W & 5 & 85.9  \\
      \cdashline{1-4}
      \AlgoName &C,H & 5 & \textbf{87.3} \\
      \bottomrule
      
    \end{tabular}
\egroup
\tiny

\end{wraptable}

\section{Ablation Study}
\label{sec:ablation}
In this section, we conduct a series of experiments to further understand the performance improvements and analyze the impact of our key contributions. For this purpose, we adopt the IAM dataset, and the baseline model is the SCATTER architecture~\cite{Litman_2020_CVPR}. Similar to our previous experiments, implementation details are described in \cref{app:implementation_det}.

We begin by exploring the different variants of AdaIN, demonstrating that our method significantly outperforms other variants. Then, we show the compatibility of our method with other augmentation pipelines demonstrating its unique benefit as a complimentary method. To analyze the performance as a function of the granularity level, we then measure the accuracy while varying the number of windows. Subsequently, we demonstrate the reliance of text recognizers on local feature statistics by increasing the hyperparameter $p$. Lastly, we perform an analysis of our method's robustness towards challenging testing conditions, demonstrating its effectiveness on relieving the model of this shortcut.

\begin{figure*}[t]
  \centering
  \includegraphics[width=0.8\textwidth]{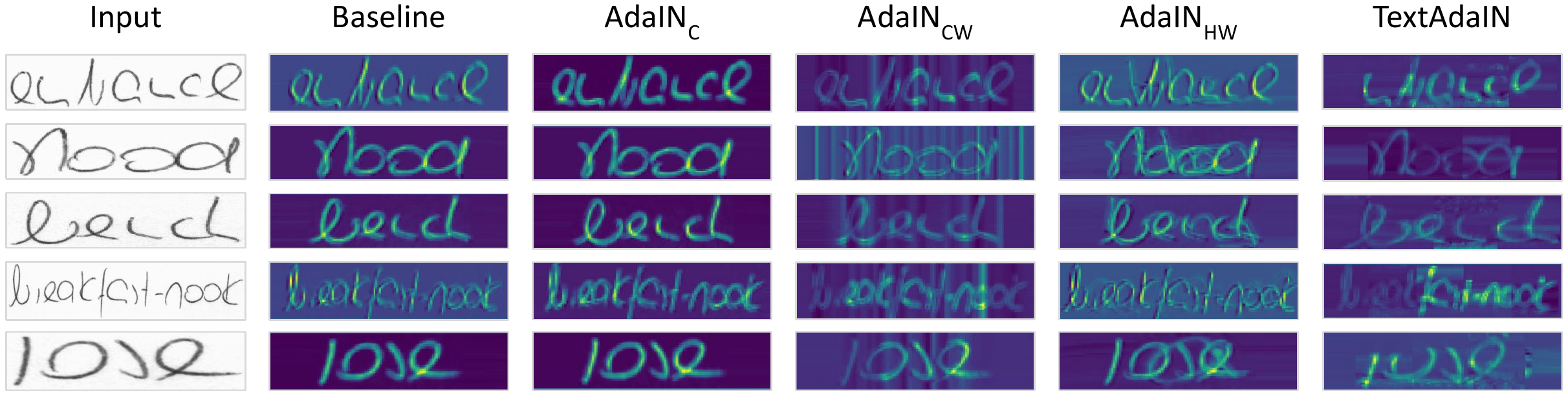}
  \caption{{\bf Information encompassed in each dimension.} Intensity maps of the features after the first convolution layer of the encoder are visualized. Each dimension of the feature space represents different information, hence, operating on different dimensions can influence the model accordingly.  \AlgoName has two noticeable effects: (1) injecting background distortions drawn from an induced distribution and (2) introducing masking on a local scale. Both regulate the reliance on local feature statistics, as further investigated in \cref{sec:corruption}}
  \label{fig:adain_config}
\end{figure*}

\subsection{AdaIN Variations}

Each dimension of the feature space represents different information about the input. Hence, modifying AdaIN to operate across different dimensions can influence the model accordingly. Our method is employed over both the channel and height dimensions between different elements of samples in a batch. Each element consists of a pre-defined number of consecutive frames in the sequential feature map. As seen in \cref{tab:spAdaIN_configuration}, \AlgoName significantly improves recognition performance as opposed to all other AdaIN variations.

To better understand the information encompassed in each of the dimensions, we visualize the feature maps of a trained baseline model in \cref{fig:adain_config}. For this purpose, we apply PCA on the spatial dimensions of the first convolutional layer's output, thus obtaining an $H\times W \times 1$ intensity map. An image depicting the spatial intensities is displayed after normalization.

As depicted in \cref{fig:adain_config}, applying $\textrm{AdaIN}_{c}$, the vanilla variation, has almost no effect on the learned features. This is in alignment with the quantitative results indicating that text recognizers are relatively invariant to changes in global statistics. As for $\textrm{AdaIN}_{c,w}$, modifying individual vertical frames introduces subtle changes to the feature map. The network can easily compensate for these distortions, leading to minimal impact on the training process.
Interestingly, applying $\textrm{AdaIN}_{h,w}$ injects text from the donor image into the feature map. This phenomenon originates from shifting each corresponding spatial location in the representation space. Clearly, without the modification of the labels, this effect will not improve the performance.

\AlgoName has two major effects, as visualized in \cref{fig:adain_config}.
The first is injecting local perturbations drawn from an induced distribution into the feature space. The distribution is induced from the manner in which \AlgoName operates, providing a correct balance between the coarse to fine distortion level. Namely, \AlgoNameNoSpace's impact is more local than $\textrm{AdaIN}_{c}$, yet more global in the sequence dimension than $\textrm{AdaIN}_{c,w}$. Therefore, the impact aligns with both the nature of the data and sequence-to-sequence approaches.
\setlength{\intextsep}{-0.4cm}%
\begin{table}
\setlength\tabcolsep{4pt}
\begin{center}
    \vspace{-0.4cm}
\bgroup
\def\arraystretch{1.1}
\centering

  \begin{center}
    \caption{\textbf{\AlgoNameNoSpace's contribution is complimentary to data augmentations effect.} Word accuracy for augmentation pipelines with (\ding{55}) and without (\checkmark) \AlgoNameNoSpace}
        \begin{tabular}{@{}llcccc@{}}
            \toprule
            \multirow{2}{*}{\textbf{Architecture}} & \multirow{2}{*}{\textbf{Augmentation Pipeline}} & \multicolumn{2}{c}{\textbf{IAM}} & \multicolumn{2}{c}{\textbf{RIMES}} \\
            & & \ding{55} & \checkmark & \ding{55} & \checkmark \\
             \midrule
            \multirow{5}{*}{SCATTER~\cite{Litman_2020_CVPR}} & \ding{55} & 84.6 & \textbf{86.5} & 91.6 & \textbf{94.2} \\
             & SCATTER~\cite{Litman_2020_CVPR} & 85.7 & \textbf{87.3} & 93.3 & \textbf{94.4} \\
             & VisionLAN~\cite{wang2021two}, ABINet~\cite{fang2021read}& 86.3 & \textbf{87.3} & 93.4 & \textbf{94.7}  \\
             & RandAug~\cite{cubuk2020randaugment} & 86.4 & \textbf{87.0} & 93.5 & \textbf{94.4}  \\
             & Luo \etal~\cite{luo2020learn} & 85.7 & \textbf{87.2} & 93.5 & \textbf{94.7}  \\
            \bottomrule
        \end{tabular}
        \tiny
  \captionsetup[table]{skip=2pt}
    \label{tab:aug_tab}
  \end{center}
\egroup
\end{center}
\end{table}

Occasionally, statistics of smooth areas (without text) are injected into regions of the feature space which represent text. This generates the second effect of local masking, in which part of the textual features undergo masking. We hypothesize that this forces the model to rely on semi-semantic information, which compensates for the missing visual cues. This was partially observed by Aberdam \etal~\cite{aberdam2020sequence} while applying horizontal cropping and in the context of speech recognition by Baevski \etal~\cite{baevski2020wav2vec}. As this analysis was performed on the feature space, we also show the influence of \AlgoName on the input space in \cref{sec:corruption}. Moreover, to see the importance of the induced distribution causing these effects, see \cref{app:induced_distribution}.

\subsection{Compatibility with Augmentation Strategies}

Similarly to \AlgoNameNoSpace, data augmentation strategies can be used to relieve a model's propensity to shortcuts and increase their robustness. Hence, they have a similar effect, yet they operate at different levels. One at the input space and the other at the feature space. To validate the compatibility of our method with augmentation strategies, we apply a variety of augmentation strategies on top of the same architecture \cite{Litman_2020_CVPR} with and without \AlgoNameNoSpace. We choose augmentation strategies ranging from other text recognition frameworks \cite{wang2021two, fang2021read}, pseudo character-level augmentations \cite{luo2020learn} and state-of-the-art augmentation pipelines \cite{cubuk2020randaugment}. \cref{tab:aug_tab} displays the word accuracy for each augmentation pipeline with and without \AlgoNameNoSpace. Applying \AlgoName in conjunction with other augmentation strategies consistently improves performance and thus the contribution of \AlgoName is complimentary to the chosen augmentation strategy. Furthermore, applying \AlgoName without any augmentation outperforms all other independent augmentation pipelines, indicating the effectiveness and impact of our method over the most common regularization technique.

\newcommand\myfontsize{\fontsize{8.5pt}{8pt}\selectfont}
\begin{table*}
\setlength\tabcolsep{2.4pt}
\begin{center}
\bgroup

  \captionsetup[table]{skip=5pt}
    \caption{\textbf{Reducing the gap in challenging testing conditions.} To demonstrate that \AlgoName indeed regulates the usage of shortcuts, we evaluate it on corrupted versions of IAM. The corruptions are divided into three different categories based on their impact type. The normalized gap provides evidence for improved robustness, especially on local corruptions}
\centering
  \begin{center}
    \myfontsize
    \begin{tabular}{l | c | c c | c c c | c c}
      \toprule
      \multicolumn{1}{c}{} & \multicolumn{1}{c}{} & \multicolumn{2}{c}{\textbf{Local Masking}} & \multicolumn{3}{c}{\textbf{Pixel-wise Distortions}} &  \multicolumn{2}{c}{\textbf{Geometric}} \\ 
      \multirow{2}{*}{Corruption Type} & \multirow{2}{*}{None} & \multirow{2}{*}{Dropout} & \multirow{2}{*}{Cutout} & Additive & Elastic & Motion & Shear & \multirow{2}{*}{Perspective} \\
      & & & & Gaussian Noise & Transform & Blur & \& Rotate & \\
      \midrule
      Baseline & 85.7 & 51.4 & 53.7 & 62.0 & 66.7 & 70.5 & 76.8 & 74.0\\

      \AlgoName  & 87.3 & 57.7 & 59.6 & 72.0 & 72.4 & 73.6 & 78.9 & 75.2 \\
    \midrule
    Gap & \color{OliveGreen}{\textbf{+1.6}} & \color{OliveGreen}{\textbf{+6.3}} & \color{OliveGreen}{\textbf{+5.9}} & \color{OliveGreen}{\textbf{+10}} & \color{OliveGreen}{\textbf{+5.7}} & \color{OliveGreen}{\textbf{+3.1}} & \color{OliveGreen}{\textbf{+2.1}} & \color{OliveGreen}{\textbf{+1.2}} \\
    Normalized Gap & 1.00 & 3.94 & 3.69 & 6.25 & 3.56 & 1.94 & 1.31 & 0.75 \\
      \bottomrule
    \end{tabular}

    \label{tab:corruption}
  \end{center}
\egroup
\end{center}
\end{table*}

\subsection{Number of Windows}

\AlgoName splits the feature map into windows along the width axis. Each window is perceived as an individual element in the AdaIN operation. As the features vary in size at different layers, we define $K$ to represent the number of elements created per sample. Thus, $K$ determines the window size at each layer. Modifying $K$ has several different effects. For example, it controls the granularity level in which the statistics are calculated and modified and the number of donors.
Therefore, an optimal value of $K$ can be found to balance the different effects. In \cref{app:num_of_windows}, we plot the performance as a function of $K$. The best result is achieved when using $K=5$. We note that the average length of English words is 4.7 characters. Thus, when $K=5$, the statistics are normalized per character on average.

\subsection{Challenging Testing Conditions}
\label{sec:corruption}

Notice that our approach leads to a similar boost on each of the datasets across all tested architectures and augmentation settings (for example $\approx +1\%$ on IAM). This suggests that this gap indeed stems from the shortcuts in the data and is not model dependent. Thus, we expect our method to perform similarly even on future and better-performing architectures. Moreover, we remind that one can detect shortcut learning at test time, given an image that violates the shortcut hypothesis (\emph{i.e.,} an ``f'' with a different level of curvature than in the train). Thus, there are potentially even more undiscovered shortcuts than our improvements on standard test benchmarks reveal. To better demonstrate the existence of unrevealed shortcut learning, we perturb the test data in a way that changes its local style without modifying the semantic content. If the models are independent of the local style, one would expect that they will be robust to such modifications.
Nevertheless, the significant performance drop (\cref{tab:curr_small}) reveals that these models are highly prone to shortcut learning, much more than discovered on the unperturbed test sets.
To demonstrate that \AlgoName indeed improves robustness towards challenging testing conditions, in \cref{tab:corruption}, we evaluate its performance on several types of corruptions, comparing it to the baseline model.

The corruptions are divided into three categories: (1) local masking, (2) pixel-wise methods and (3) geometric transformations. For each corruption, we display the gap between the performance of the baseline versus the \AlgoName model. To accentuate the improvement provided by our method, we also display the normalized gap - the ratio between the gap on the corrupted data and the gap on the original data. This normalization removes the performance advantage of \AlgoName on the original data. The results indicate that \AlgoName improves the model's robustness towards \textit{o.o.d} testing scenario. For example, the gap on additive noise is 10\%.
We note that the normalized gap, which represents \AlgoNameNoSpace's robustness gains, is substantially higher in local-based corruptions rather than geometric corruptions which are applied globally. This provides further evidence that TextAdaIN regulates the reliance on local statistics in text recognizers. Despite the benefits of our method, \cref{tab:corruption} indicates that there are potentially even more undiscovered shortcuts as a significant gap exists when comparing to the original unperturbed test set.

\subsection{Reliance on Local Statistics}
\begin{wraptable}[20]{r}{5cm}
\normalsize
\begin{center}
\small
\caption{\textbf{Probability of applying \AlgoNameNoSpace.} Applying AdaIN at high values of $p$ has little to no effect on the performance. In contrast, profusely applying \AlgoName distorts important information that the model relies on. 
}
\bgroup
        \begin{tabular}{lccc}
        \toprule
        Method & $p$ & IAM Acc. \\
        \midrule
        Baseline & -- & 85.7\\
        \midrule
        \multirow{3}{*}{AdaIN} & 0.01 & 85.4\\
        & 0.1 & 85.9 \\
        & 0.25 & 85.8 \\
        \midrule
        \multirow{6}{*}{\AlgoName} & 0.001 & 86.1 \\
         & 0.01 & \textbf{87.3} \\
         & 0.05 & 86.8 \\
         & 0.1 & 78.0 \\
         & 0.25 & 74.7 \\
        \bottomrule
    \end{tabular}
\egroup
\tiny

\label{tab:spadain_vs_padain}
\vspace{-0.85cm}
\end{center}
\end{wraptable}

\label{sec:reliance}
In this subsection, we wish to further assert text recognizers' reliance on local statistics. Nuriel et al. \cite{nuriel2021permuted} observed that applying AdaIN at high values of $p$ resulted in a significant degradation of classification performance. We can thus infer that image classifiers depend on global statistics.
For text recognizers, as shown in \cref{tab:spadain_vs_padain}, this is not the case. 
Increasing the value of $p$, when applying AdaIN, only slightly affects the results. This indicates that global statistics are less significant in text recognition.
In contrast, applying \AlgoName with a high value of $p$ decreases performance substantially. This implies that profusely applying \AlgoName distorts important information that the model relies on. Therefore, text recognizers are prone to develop an unintended shortcut solution in the form of local statistics. If applied correctly, with the right $p$, \AlgoName can alleviate this shortcut.

\section{Conclusion}


Text recognizers leverage convolutional layers to extract rich visual features, and hence are extremely powerful. However, in this work, we expose their propensity towards learning an unintended ``shortcut'' strategy, whereby they overly rely on local statistics. Consequently, exhibiting a sensitivity towards subtle modifications that preserve image content. To relieve text recognizers' shortcut learning, we introduce \AlgoNameNoSpace, a normalization-based method which distorts the feature space in a local manner and effectively regulates the reliance on local statistics.
Our method achieves state-of-the-art results on handwritten text recognition benchmarks and improves robustness towards challenging testing conditions.
\AlgoName is also applicable to various recognition architectures and to the domain of scene text images. Furthermore, it can be implemented simply in a few lines of code and effortlessly integrated into a mini-batch training procedure.

By taking into account the nature of the data and the archictectural characteristics, this shortcut was exposed. Yet, as we have seen, the prevalence of shortcuts is still at large. As future work, we wish to explore other shortcuts pertaining to text recognizers.


\clearpage
%
%
\bibliographystyle{splncs04}
\bibliography{egbib}

\clearpage

\setlength{\intextsep}{0.4cm}%

\appendix
 
\section{SCATTER - Baseline Architecture}
\label{app:scatter}
As mentioned in the paper, throughout this work our baseline architecture is a state-of-the-art recognizer named SCATTER~\cite{Litman_2020_CVPR}. 
The SCATTER architecture consists of four main components:
\begin{enumerate}[nolistsep]
\item Transformation: A rectification module that aligns the input text
image using a Thin Plate Spline (TPS) transformation \cite{Liu2016STARNetAS,shi2016robust}.
\item Feature extraction: A convolutional neural network
(CNN) that extracts features from the rectified image. Similar to~\cite{cheng2017focusing,Baek2019clova,Litman_2020_CVPR,aberdam2020sequence,aberdam2022multimodal}, a 29-layer ResNet is employed as the backbone. Subsequently, the features are mapped to a sequence of frames, denoted by $V = [v_1, v_2, ..., v_T]$, each corresponding to different receptive fields in the image.
\item Visual feature refinement: An intermediate supervision in the form of a CTC decoder \cite{Graves2006ctc} is employed to provide direct supervision for each frame in the visual features $V$.
\item Selective-Contextual Refinement Block: Contextual features are extracted using a two-layer bi-directional LSTM encoder. These features are concatenated to the visual features, $V$. The features are then fed into a selective decoder and into a subsequent block if it exists.
These blocks can be stacked together to improve results. In this work, for convenience, we set the number of blocks to two.
\end{enumerate}

\section{Image Corruptions}

\label{app:corruption}
\begin{figure*}[t]
 \centering
	\includegraphics[width=\textwidth]{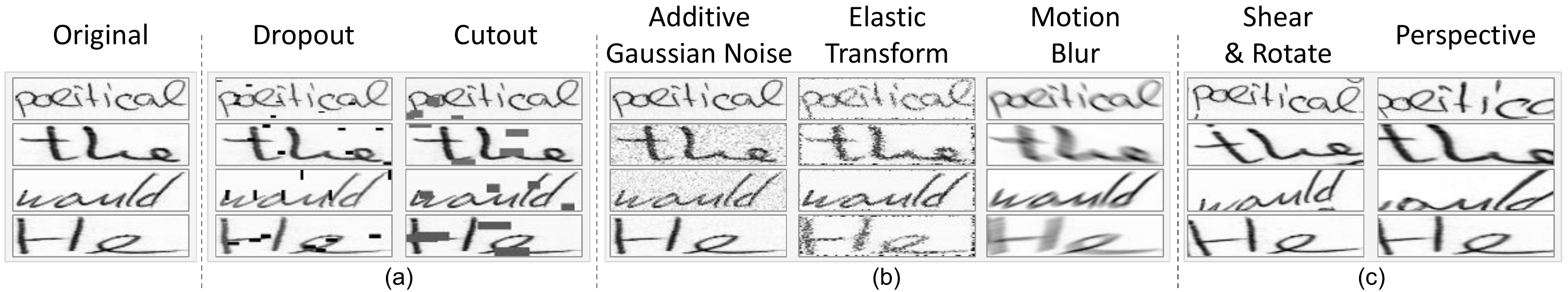}
	\caption{\label{fig:corruptions}
		{\bf Visualization of selected corruptions}. To assert \AlgoNameNoSpace's advantage on challenging testing conditions, we compare its performance to the baseline while applying different types of corruptions. We visualize the different corruptions and divide them into categories based on their impact: local masking, pixel-wise distortions and geometric.
	}
\end{figure*}

In Sections 3.1 and 5.4, we compare the performance of the baseline model and the \AlgoName version on corrupted versions of the IAM test set. \cref{fig:corruptions} contains original images from the test set on the first column and the results of applying each of the corruptions on the following columns. They are split into three categories based on their impact: (a) local masking, (b) pixel-wise distortions and (c) geometric. The corruptions are applied utilizing the \textit{imgaug}~\cite{imgaug} package as explicitly written below.

\begin{lstlisting}[language=Python]
from imgaug import augmenters as iaa
corruptions = {
    'original': iaa.Noop(),
    'dropout': iaa.CoarseDropout((0.0, 0.05),
    'cutout': iaa.Cutout(nb_iterations=4),
    'additiive_gaussian_noise': iaa.AdditiveGaussianNoise(scale=(0, 0.2*255)),
    'elastic_transformation':  iaa.ElasticTransformation(alpha=(0, 5.0), sigma=0.5),
    'motion_blur': iaa.MotionBlur(k=15, angle=[-45, 45]),
    'shear_rotate': iaa.Affine(shear=(-10, 10),rotate=(-10, 10), mode='reflect'),
 size_percent=(0.02, 0.25)),
    'perspective': iaa.PerspectiveTransform(scale=(0.05, 0.2)),
}
\end{lstlisting}

\section{Datasets}
\label{app:dtatsets}

\begin{figure}[t]
 \centering
	\includegraphics[width=0.9\columnwidth]{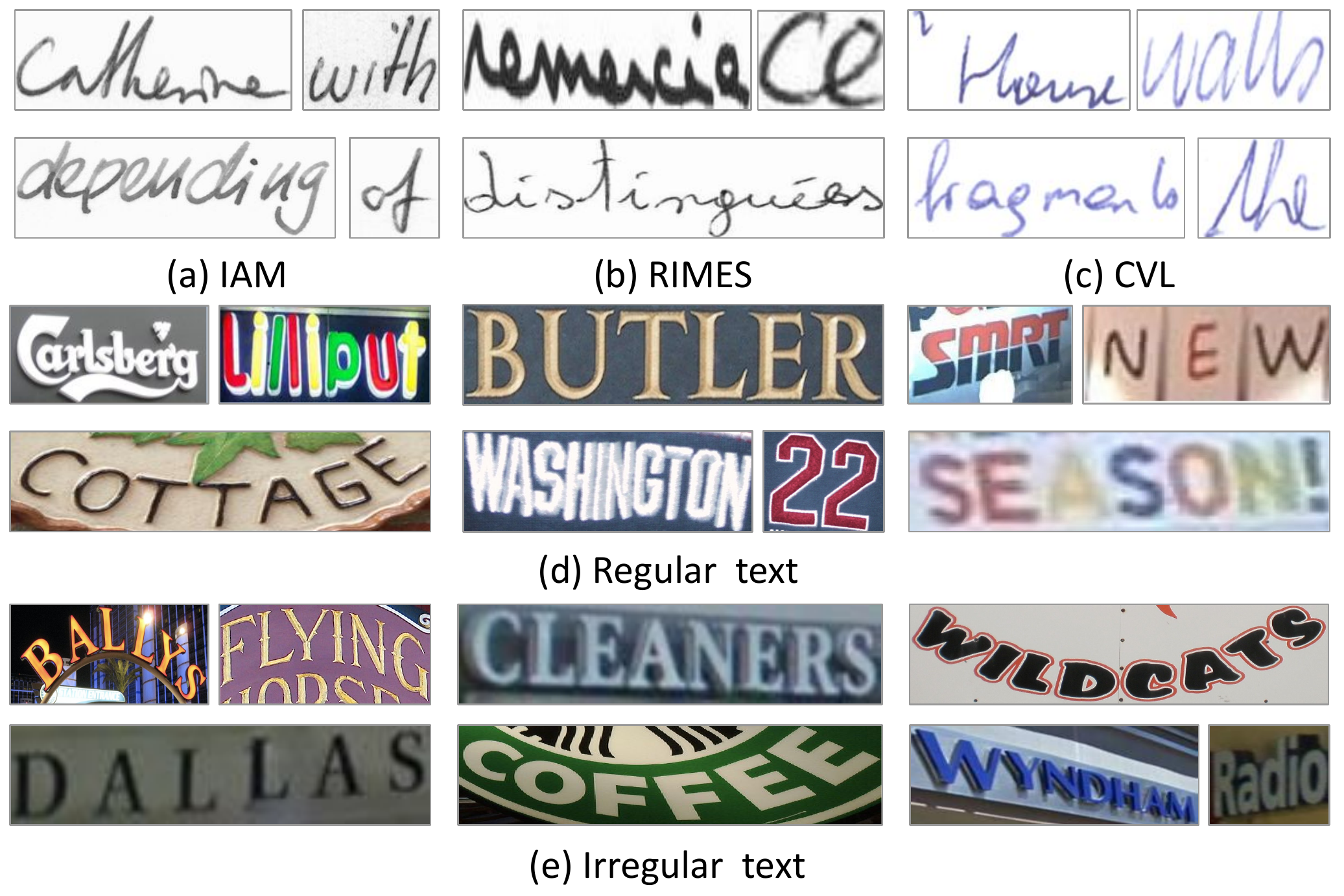}
	\caption{\label{fig:dataset_samples} \textbf{Dataset samples.} 
Examples of images from each dataset. 
	}
	\vspace{-0.5cm}
\end{figure}

In this work, we consider the following public datasets for handwriting and scene text. Samples from the different datasets are depicted in \cref{fig:dataset_samples}. 

\subsection{Handwritten text}
For handwriting recognition, we consider three datasets: 
\begin{itemize}[noitemsep]
    \item \textbf{IAM} \cite{marti2002iam} handwritten English text dataset, written by 657 different writers. This dataset contains 101,400 correctly segmented words, partitioned into writer independent training, validation and test.
    \item \textbf{CVL} \cite{kleber2013cvl} handwritten English text dataset, written by 310 different writers. 27 of the writers wrote 7 texts, and the other 283 writers wrote 5 texts. This dataset contains 84,990 correctly segmented words, partitioned into writer independent training and test. We use the same partitions as in \cite{fogel2020scrabblegan, aberdam2020sequence}, in which the 310 writers are used for training and the additional 27 writers are considered test.
    \item \textbf{RIMES} \cite{grosicki2009icdar} handwritten French text dataset, written by 1300 different writers. This dataset contains 66,480 correctly segmented words, partitioned into training, validation and test sets that are independent of writers.
\end{itemize}

\subsection{Scene text}
For training the scene text models, we utilized only synthetic datasets:
\begin{itemize}[noitemsep]
    \item {\textbf{MJSynth}} (MJ) \cite{Zisserman2014mj} a synthetic text in image dataset which contains 9 million word-box images, generated from a lexicon of 90K English words.
    \item \textbf{SynthText} (ST) \cite{Zisserman206st} a synthetic text in image dataset, containing 5.5 million words, designed for scene text detection and recognition.
    \item \textbf{SynthAdd} (SA) \cite{Wang2019sar} only when training SCATTER for scene text, as in the original paper~\cite{Litman_2020_CVPR}, SA was also utilized for training data. This dataset was generated using the same synthetic engine as in ST and contains 1.2 million word box images. SA is used for compensating the lack of non-alphanumeric characters in the training data.
\end{itemize}

Aligned with many scene text recognition manuscripts (e.g. \cite{Bai2018aster, Baek2019clova, Wang2019sar, Litman_2020_CVPR}), we evaluate our models using seven scene text datasets: ICDAR2003, ICDAR2013, IIIT5K, SVT, ICDAR2015, SVTP and CUTE. Those datasets are commonly divided into regular and irregular text according to the text layout.
\newline
\textbf{Regular text} datasets are composed of:
\begin{itemize}[noitemsep]
    \item \textbf{IIIT5K}~\cite{Mishra2012sj} consists of 2000 training and 3000 testing images that are cropped from Google image searches.
    \item \textbf{SVT}~\cite{Wang2011bottom} is a dataset collected from Google Street View images and contains 257 training and 647 testing cropped word-box images.
    \item \textbf{ICDAR2003}~\cite{Lucas2003ic03} contains 867 cropped word-box images for testing.
    \item \textbf{ICDAR2013}~\cite{Karatzas2013ic13} contains 848 training and 1,015 testing cropped word-box images.
\end{itemize}
\textbf{Irregular text} datasets are composed of:
\begin{itemize}[noitemsep]
    \item \textbf{ICDAR2015}~\cite{Karatzas2015ic15} contains 4,468 training and 2,077 testing cropped word-box images, all captured by Google Glass, without careful positioning or focusing.
    \item \textbf{SVTP}~\cite{Phan2013svtp} is a dataset collected from Google Street View images and consists of 645 cropped word-box images for testing.
    \item \textbf{CUTE 80}~\cite{Risnumawan2014cute} contains 288 cropped word-box images for testing, many of which are curved text images. 
\end{itemize}

\section{Implementation Details}
\label{app:implementation_det}
In our experiments, we utilize four types of architectures. The first is SCATTER, the second and third are two other are variants of the Baek~\etal\cite{Baek2019clova} framework and the last is AbiNet \cite{fang2021read}. All models are trained and tested using the PyTorch
framework on a Tesla V100 GPU with 16GB memory.

We follow the training procedure performed in \cite{Litman_2020_CVPR}. Accordingly, models are trained using the AdaDelta optimizer with: a decay rate of 0.95, gradient clipping with a magnitude of 5 and batch size of 128. During training, 40\% of the input images are augmented by randomly resizing them and adding extra distortion. Models trained on handwriting and scene text datasets are trained for 200k iterations and 600k iterations, respectively. Model selection is performed on the validation set, which for scene text is the union of the IC13, IC15, IIIT5K and SVT training splits and for handwriting is the predefined validation sets. When utilizing SCATTER, all images are resized to $32\times 128$ and are in RGB format. For evaluation, word accuracy measured is case insensitive. We refer the reader for any additional implementation details, both during inference and training, to the original papers \cite{Baek2019clova, Litman_2020_CVPR}.

To reproduce the results for AbiNet we downloaded the pretrained models from the official open source implementation (\url{https://github.com/FangShancheng/ABINet}) and re-ran the fine-tune step. When adding \AlgoNameNoSpace, we apply it in both phases, in the pre-training step as well as in the fine-tuning step. We use the default configurations as defined in the repository. For handwriting, we use the default configurations but run for 100 epochs and adjust the learning rate schedule to 60, 25, 15.

For \AlgoNameNoSpace, we use $p=0.01$ and split images into $K=5$ windows. In the cases where the number of windows does not divide the width, we use the maximum width that can be divided and ignore the remainder.

\section{Failure Cases}
\label{app:fail_cases}
\begin{figure*}
  \centering
  \includegraphics[width=0.95\columnwidth]{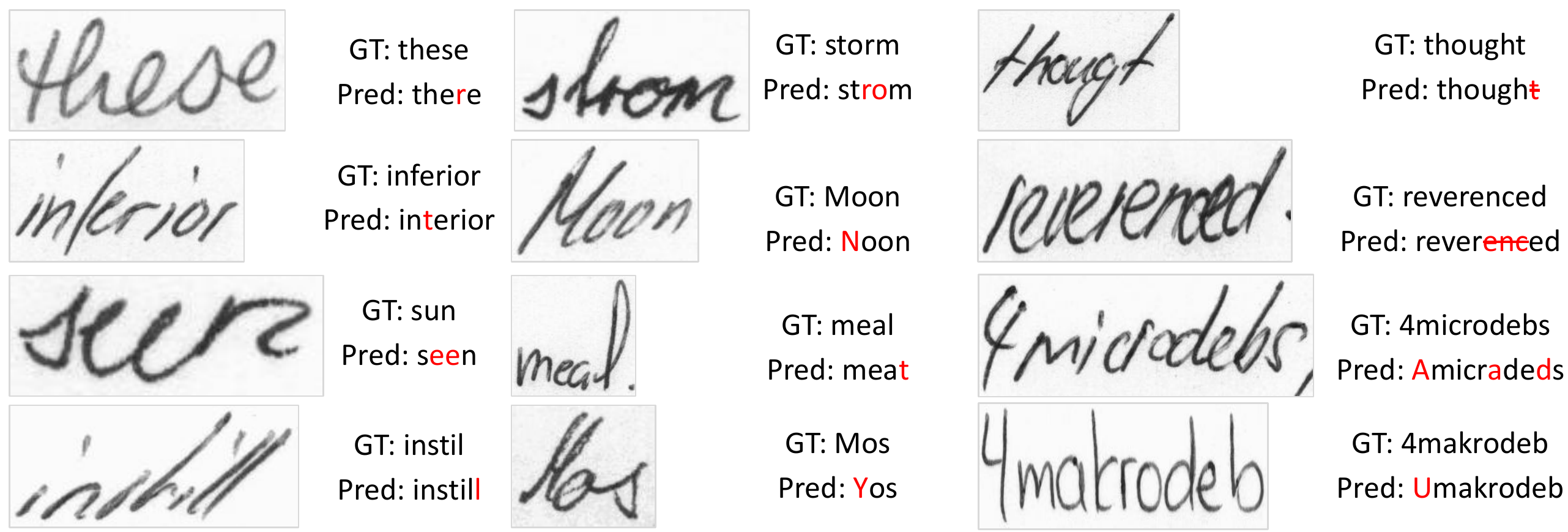}
  \caption{{\bf Failure cases.} Samples of failure cases on the IAM dataset. GT stands for the ground truth annotation, and Pred is the predicted result. Prediction errors are marked in red, and missing characters are annotated by strike-through.}
  \label{fig:fail_case}
\end{figure*}

In \cref{fig:fail_case}, we display failure cases of our method on the IAM dataset. The failure cases are mostly composed of highly cursive text, unclear handwriting styles and ambiguous cases. Adding additional context by employing a line-level approach can assist in rectifying these types of errors, nevertheless, line-level recognition has its own caveats.

\section{The importance of an induced distribution}
\label{app:induced_distribution}

We show the importance of sampling from an induced distribution, specifically the distribution derived by the representation spaces of natural text images. As shown in \cref{tab:textadain_vs_noise}, using Gaussian noise or background images only slightly increases performances. In contrast, \AlgoNameNoSpace, which samples from the appropriate induced distribution, shows the highest increase in performance.

\begin{table*}
\begin{center}
\small
\bgroup
\caption{\textbf{The importance of the induced distribution.} Best performance is achieved only when injecting distortions from an induced distribution namely, other text images}
\label{tab:textadain_vs_noise}
        \begin{tabular}{lcc}
        \toprule
        Injection Method & Donors & Accuracy \\
        \midrule
        Baseline & X & 85.7\\
        
        \midrule
        \AlgoName & Gaussian Noise & 86.1 \\
        \AlgoName & Blank Image & 86.2 \\
        \midrule
        \AlgoName & Text Images & \textbf{87.3}\\
        \bottomrule
    \end{tabular}
\egroup
\tiny

\end{center}
\end{table*}

\section{Number of Windows}
\label{app:num_of_windows}
\begin{figure}[t]
  \centering
  \includegraphics[width=0.6\columnwidth]{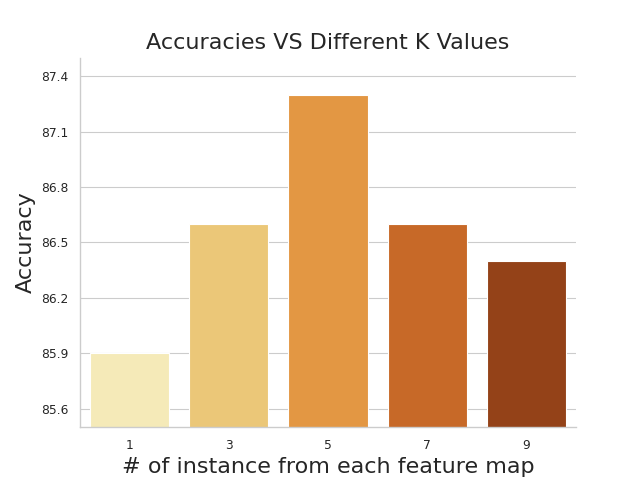}
  \caption{{\bf Number of windows.} Varying the number of windows extracted per sample has multiple effects, including the granularity level and the number of donor images.}
  \label{fig:window_size}
\end{figure}

As mentioned in the main paper, \AlgoName splits the feature map into windows along the width axis. As the features vary in size at different layers, we define $K$ to represent the number of elements created per sample. Thus, $K$ determines the window size at each layer. Modifying $K$ has several different effects. For example, it controls the granularity level in which the statistics are calculated and modified and the number of donors.
Therefore, an optimal value of $K$ can be found to balance the different effects. In \cref{fig:window_size}, we plot the performance as a function of $K$. The best result is achieved when using $K=5$.

\section{\AlgoName Pseudo-Code}
\label{app:code}
In this section, we provide pseudo-code for \AlgoNameNoSpace. The code includes two functions not explicitly implemented: \textit{create\_windows\_from\_tensor}, \textit{revert\_windowed\_tensor}.
The first function represents the mapping: 
\begin{equation*}
    X\in \mathbb{R}^{B\times C \times H \times W} \rightarrow \widehat{X}\in \mathbb{R}^{B\cdot K \times C \times H \times \frac{W}{K}} \,,
\end{equation*} and the second represents the corresponding inverse mapping. As mentioned in the paper, we do not backpropogate through $\mu_{c,h}(\hat{x}_{\pi(i)}), \sigma_{c,h}(\hat{x}_{\pi(i)})$ and thus, detach is used.

\begin{lstlisting}[language=Python]
def TextAdaIN(x, p=0.01, k=5, eps=1e-4):
    # input x - a pytorch tensor
    if rand() > p:
        return x
    N, C, H, W = x.size()
    # split into windows
    x_hat = create_windows_from_tensor(x,k) 
    # calculate statistics
    feat_std = sqrt(x_hat.var(dim=3)+ eps)
    feat_std = feat_std.view(N*k, C, H, 1) 
    feat_mean = x_hat.mean(dim=3)
    feat_mean = feat_mean.view(N*k, C, H, 1)
    # perform permutation
    perm_indices = randperm(N*k)
    perm_feat_std = feat_std[perm].detach()
    perm_feat_mean = feat_mean[perm].detach()
    # normalize
    x_hat = (x - feat_mean) / feat_std
    # swap
    x_hat = x_hat * perm_feat_std
    x_hat += perm_feat_mean
    # merge windows
    x = revert_windowed_tensor(x_hat, k, W)
    return x
\end{lstlisting}

\end{document}